\documentclass[letterpaper]{article} 

\usepackage{aaai25}  
\usepackage{times}  
\usepackage{helvet}  
\usepackage{courier}  
\usepackage[hyphens]{url}  
\makeatletter

\usepackage{graphicx} 
\usepackage{natbib}  
\usepackage{caption} 
\usepackage{algorithm}
\usepackage{algorithmic}

\usepackage{newfloat}
\usepackage{listings}
\DeclareCaptionStyle{ruled}{labelfont=normalfont,labelsep=colon,strut=off} 
\floatstyle{ruled}
\newfloat{listing}{tb}{lst}{}
\floatname{listing}{Listing}
\makeatletter

\usepackage{multirow}
\usepackage{amsmath}
\usepackage{xspace}
\usepackage{makecell}
\usepackage{adjustbox}
\usepackage{array} 
\usepackage{amsfonts} 
\usepackage{nicefrac}
\usepackage{microtype}
\usepackage{color, colortbl}
\usepackage[dvipsnames]{xcolor}
\usepackage{pifont}
\definecolor{mygray}{gray}{.88}
\newcommand{\cmark}{\color{ForestGreen}\ding{51}}%
\newcommand{\xmark}{\color{Red}\ding{55}}%
\usepackage{parskip}
\usepackage{booktabs}
\usepackage{subcaption}
\makeatletter

\title{Adaptive Calibration: A Unified Conversion Framework \\ of Spiking Neural Network}
\author{
    Ziqing Wang\textsuperscript{\rm 1, 2}\equalcontrib,
    Yuetong Fang\textsuperscript{\rm 1}\equalcontrib, 
    Jiahang Cao\textsuperscript{\rm 1}, 
    Hongwei Ren\textsuperscript{\rm 1}, 
    Renjing Xu\textsuperscript{\rm 1}\thanks{Corresponding Author}
}
\affiliations{
    \textsuperscript{\rm 1}The Hong Kong University of Science and Technology (Guangzhou), China\\
    \textsuperscript{\rm 2}Northwestern University, USA\\
    ziqingwang2029@u.northwestern.edu, \\\{yfang870, jcao248, hren066\}@connect.hkust-gz.edu.cn, renjingxu@hkust-gz.edu.cn

}

\usepackage{bibentry}

\begin{document}

\maketitle

\begin{abstract}
Spiking Neural Networks (SNNs) are seen as an energy-efficient alternative to traditional Artificial Neural Networks (ANNs), but the performance gap remains a challenge. While this gap is narrowing through ANN-to-SNN conversion, substantial computational resources are still needed, and the energy efficiency of converted SNNs cannot be ensured. To address this, we present a unified training-free conversion framework that significantly enhances both the performance and efficiency of converted SNNs. Inspired by the biological nervous system, we propose a novel Adaptive-Firing Neuron Model (AdaFire), which dynamically adjusts firing patterns across different layers to substantially reduce the \textit{Unevenness Error} - the primary source of error of converted SNNs within limited inference timesteps. We further introduce two efficiency-enhancing techniques: the Sensitivity Spike Compression (SSC) technique for reducing spike operations, and the Input-aware Adaptive Timesteps (IAT) technique for decreasing latency. These methods collectively enable our approach to achieve state-of-the-art performance while delivering significant energy savings of up to \textbf{70.1\%}, \textbf{60.3\%}, and \textbf{43.1\%} on CIFAR-10, CIFAR-100, and ImageNet datasets, respectively. Extensive experiments across 2D, 3D, event-driven classification tasks, object detection, and segmentation tasks, demonstrate the effectiveness of our method in various domains. The code is available at: \texttt{https://github.com/bic-L/burst-ann2snn}.

\end{abstract}

%

\section{Introduction}

Spiking Neural Networks (SNNs) have gained great attention for their potential to revolutionize the computational efficiency of artificial intelligence systems. Unlike traditional Artificial Neural Networks (ANNs), which rely on the intensity of neuron activations, SNNs utilize the timing of sparse and discrete spikes to encode and process information~\cite{maass1997networks}. This spike-based computing paradigm is particularly suited to neuromorphic hardware, which utilizes spiking neurons and synapses as fundamental components~\cite{daviesLoihiNeuromorphicManycore2018, davies2021advancing, akopyanTruenorthDesignTool2015}. In SNNs, incoming spikes trigger the retrieval of synaptic weights from memory and generate subsequent spike messages routed to other cores, promoting energy-efficient operations over traditional energy-intensive matrix multiplications. The inherent properties of SNNs, coupled with recent advances in neuromorphic hardware, position them as a promising solution for developing energy-efficient and high-performance artificial intelligence systems. Consequently, recent works spanning classification~\cite{wang2023masked,zhouSpikformerWhenSpiking2022,dengTemporalEfficientTraining2022}, tracking~\cite{zhangSpikingTransformersEventBased2022}, and image generation~\cite{cao2024spiking} are striving to combine advanced network architectures~\cite{vaswaniAttentionAllYou2017, he2023reti} with this spike-driven computing paradigm.

\begin{figure}[!t]
\centering
\setlength\tabcolsep{3.5pt}
\renewcommand{\arraystretch}{0.9} 
\scriptsize
\captionof{table}{\textbf{
Comparison of the proposed method versus existing methods on ImageNet (VGG-16).} 'T' refers to averaged inference time steps, and 'Time Cost' represents the total GPU hours before final inference. Our Adaptive Calibration method achieves competitive accuracy with fewer timesteps and lower energy consumption. Notably, it only requires a short setup time and eliminates the need for re-training.}
\label{tb:comparison}
\vspace{-0.05in} 
\begin{tabular}{lccccc}
  \toprule
  \textbf{Method}      & \textbf{T} & \textbf{Acc. (\%)} & \textbf{Energy (mJ)} & \textbf{Training} & \textbf{Time Cost (h)} \\ \midrule
  \textbf{QCFS}~\textsuperscript{ICLR}        & 32         & 68.47              & 77.41                    & \cmark               & 742.51                  \\
  \textbf{FastSNN}~\textsuperscript{TPAMI}     & 7          & 72.95              & 16.93                    & \cmark               & 484.45                  \\
  \textbf{Calibration}~\textsuperscript{ICML} & 32         & 62.14              & 57.13                     & \xmark                &   0.06               \\
  \textbf{Ours}        & 5.72          & 73.46              & 22.47                   & \xmark                 & 0.09                  \\
  \bottomrule
\end{tabular}
\vspace{-0.1in} 
\end{figure}

Despite the significant energy efficiency benefits offered by SNNs, achieving performance on par with ANNs remains a challenge. The main difficulty stems from the non-differentiability of discrete spikes and the complex computational graph due to multi-timestep operations. These factors make training SNNs from scratch both intricate and computationally demanding~\cite{neftciSurrogateGradientLearning2019}. To address this challenge, ANN-to-SNN conversion techniques have emerged as a promising approach, enabling the direct conversion of pre-trained ANNs into high-performance SNNs. Recent advancements aim to reduce conversion errors by replacing the ReLU activation in ANNs with specially designed quantized functions, following extra training.  Typically, the quantized ReLU activation function is employed, as it better mimics spiking neuron dynamics~\cite{stocklOptimizedSpikingNeurons2021, buOptimalANNSNNConversion2021, dingOptimalAnnsnnConversion2021}. However, these re-training-based methods come with certain disadvantages. Firstly, as shown in Tab.~\ref{tb:comparison}, these studies require training an intermediate surrogate ANN on top of the original ReLU-based ANN, extending the overall training period.
Secondly, they compromise the inherent energy efficiency of SNNs by necessitating longer simulation timesteps to minimize conversion errors between quantized ReLU and spiking neurons during inference, thus increasing synaptic operations.

\begin{figure}[!t]
\centering
\includegraphics[width=0.8\columnwidth]{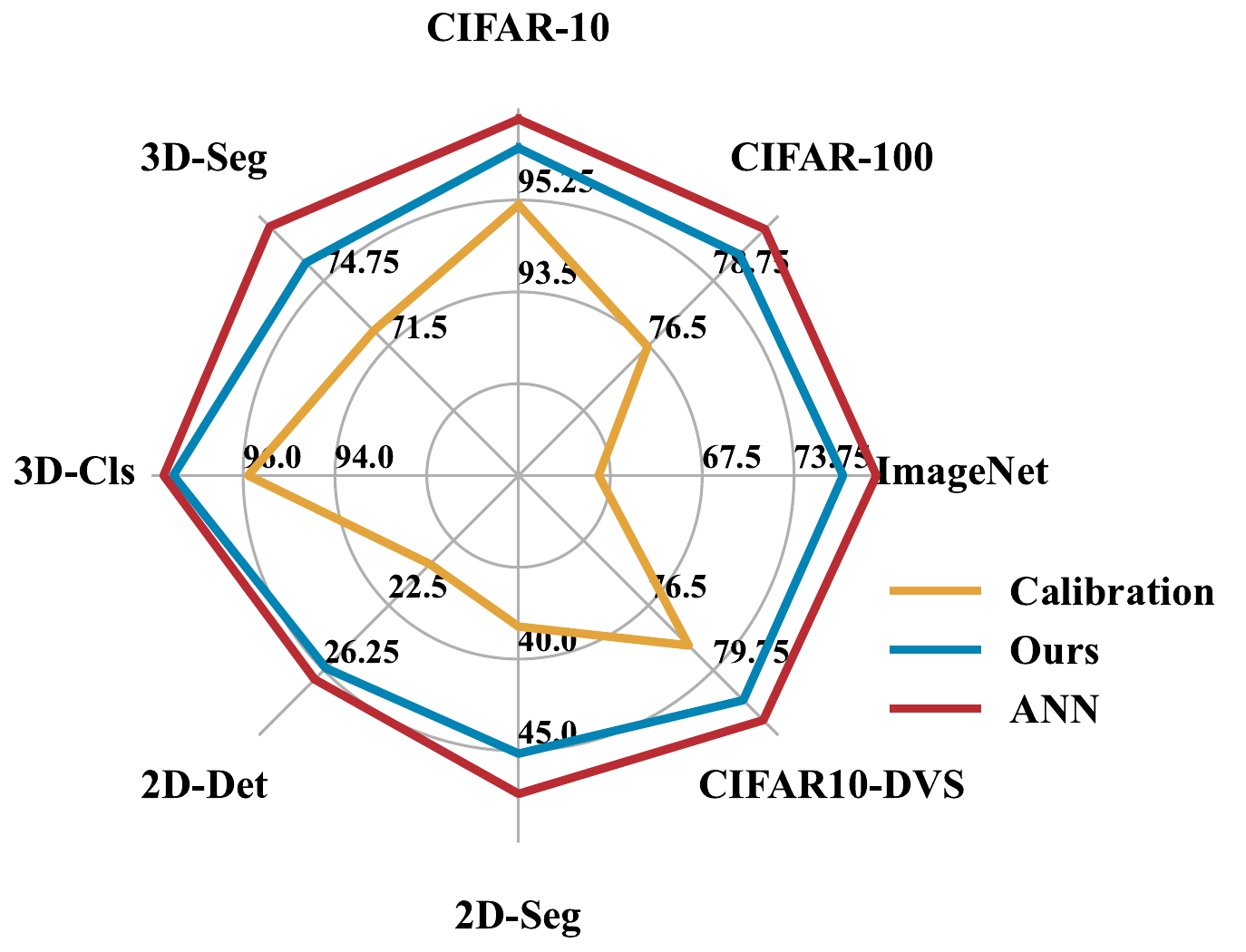}
\vspace{-0.05in} 
\caption{\textbf{Performance comparison on different tasks.} }
\vspace{-0.2in}
\label{fig:radar}
\end{figure}

SNN Calibration~\cite{liFreeLunchANN2021} offers a fast, training-free alternative to re-training-based methods for converting advanced ANN architectures into spike-driven models, completing in minutes without extra training, as shown in Table~\ref{tb:comparison}. However, the previous study focuses solely on minimizing errors in the network output space, neglecting the differences between ANN and SNN neurons, such as unevenness errors. This results in a larger performance gap, especially under the same inference timestep, limiting practical use in real-time, energy-constrained applications. To address this issue, we draw inspiration from the burst-firing mechanism, widely observed in the human brain. This mechanism features rapid sequences of action potentials that have proven to enhance the reliability of information transmission, in contrast to regular-spiking cells that fire at consistent rates~\cite{connorsIntrinsicFiringPatterns1990, izhikevichBurstsUnitNeural2003, lismanBurstsUnitNeural1997}. 
Remarkably, this mechanism 
is well-supported by neuromorphic hardware, such as Intel's Loihi 2 and Synsense's Speck ~\cite{orchard2021efficient, daviesLoihiNeuromorphicManycore2018, davies2021advancing, akopyanTruenorthDesignTool2015}. 

\begin{figure*}[!t]
\centering
\includegraphics[width=0.95\linewidth]{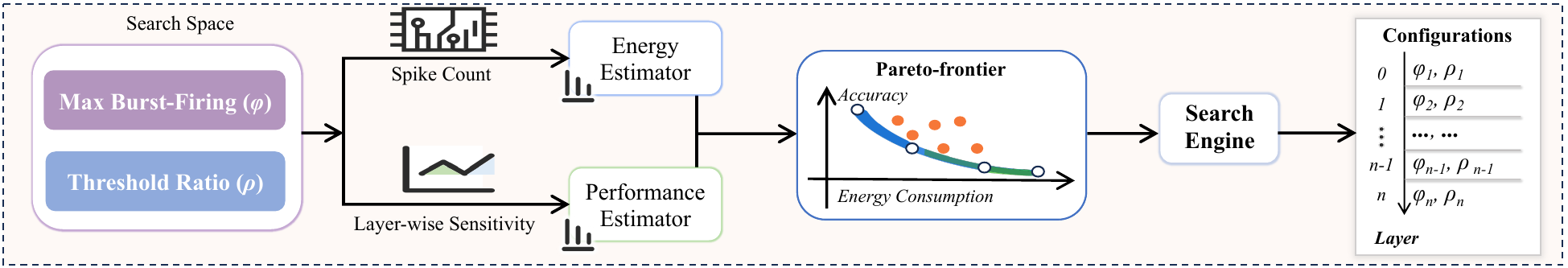}
\vskip -0.05in
\caption{\textbf{Optimization Process for Adaptive Calibration.} The process begins within a search space containing candidates for Max Burst-firing Patterns ($\varphi$) and Threshold Ratio ($\rho$). The estimators assess these candidates to evaluate their performance and energy efficiency. Optimum configurations of each layer are then selected using the Pareto-frontier method.}
\vskip -0.1in
\label{mainfig}
\end{figure*}

This motivates us to investigate an Adaptive Calibration framework that utilizes adaptable firing patterns of spiking neurons across layers, aiming to simultaneously reduce ANN-to-SNN conversion error, latency, and energy consumption, as shown in Fig~\ref{mainfig}. In summary, the contributions of our paper are as follows:
\begin{itemize}
    \item An Adaptive-Firing Neuron Model (AdaFire) integrated into the SNN Calibration process to automatically search for optimum firing patterns, significantly reducing Unevenness Error—the primary error source during conversion within limited timesteps.
    \vspace{-0.02in}
    \item A Sensitivity Spike Compression (SSC) technique that dynamically adjusts thresholds based on layer sensitivity, ensuring energy efficiency of the converted SNNs.
    \vspace{-0.02in}
    \item An Input-aware Adaptive Timesteps (IAT) technique that adjusts timesteps based on input complexity, further decreasing energy consumption and latency.
    \vspace{-0.02in}
    \item Extensive experiments across multiple domains, demonstrating state-of-the-art performance and remarkable energy savings up to \textbf{70.1\%}, \textbf{60.3\%}, and \textbf{43.1\%} for CIFAR-10, CIFAR-100, and ImageNet datasets, respectively.
\end{itemize}

\section{Related Work}
\label{sec:pre}
\paragraph{Spiking Neuron Model.} In SNNs, inputs are transmitted through the neuronal units, typically the Integrate-and-Fire (IF) spiking neuron in ANN-to-SNN conversions~\cite{dingOptimalAnnsnnConversion2021,liFreeLunchANN2021a, buOptimalANNSNNConversion2021a}:
\begin{align}
\label{eq2}
&{u}^{(\ell)}(t+1)={v}^{(\ell)}(t)+{W}^{(\ell)} {s}^{(\ell)}(t) \\
&{v}^{(\ell)}(t+1)={u}^{(\ell)}(t+1)-{s}^{(\ell+1)}(t) \\
&{s}^{(\ell+1)}(t)= \begin{cases}V_{t h}^{(\ell)} & \text { if } {u}^{(\ell)}(t+1) \geq V_{t h}^{(\ell)} \\
0  & otherwise\end{cases}
\end{align}
\noindent where ${u}^{(\ell)}(t+1)$ denotes the membrane potential of neurons before spike generation, ${v}^{(\ell)}(t+1)$ denotes the membrane potential of neurons in layer $\ell$ at time step $t+1$, corresponding to the linear transformation matrix ${W}^{\ell}$, the threshold $V_{t h}^{(\ell)}$, and binary input ${s}^{\ell}(t)$ of layer $\ell$.

\paragraph{Burst-firing Neurons.} Recent studies have integrated burst-firing neurons into SNNs to more accurately mimic the intricate dynamics of biological neural systems~\cite{park2019fast, lan2023efficient, liEfficientAccurateConversion2022}. These neurons are supported by neuromorphic hardware, such as Intel's Loihi 2 and Synsense's Speck~\cite{orchard2021efficient, daviesLoihiNeuromorphicManycore2018}, enabling better performance. However, current methods often disregard the energy costs and apply burst-firing patterns uniformly across all SNN layers, missing the subtle layer-specific sensitivities and the balance between enhanced performance and increased energy consumption. To address these challenges, our research introduces the Adaptive Calibration framework that automatically optimizes firing patterns, enhancing performance and energy efficiency within limited timesteps without training.

\paragraph{ANN-to-SNN conversion and SNN Calibration} The fundamental principle of ANN-to-SNN conversion is to ensure that the converted SNN closely approximates the input-output function mapping of the original ANN:
\begin{equation}
x^{(\ell)} \approx \overline{s}^{(\ell)}=\frac{1}{T} \sum_{t=0}^T s^{(\ell)}(t)
\end{equation}
where $x^{(\ell)}$ represents the activation input of the ANN model, and $\overline{s}^{(\ell)}$ denotes the averaged binary input over $T$ timesteps in the converted SNN. It is important to note that this approximation becomes valid only as $T$ approaches infinity.

To address this limitation, prior works~\cite{hoTCLANNtoSNNConversion2021, dingOptimalAnnsnnConversion2021, buOptimalANNSNNConversion2021} proposed to replace the \textit{ReLU} activation function in the original ANNs with a trainable $Clip$ function, then find the optimal data-normalization factor through an additional training process to consider both accuracy and latency in the converted SNNs:
\begin{align}
\label{eq3}
\overline{{s}}^{(\ell+1)} &= ClipFloor\left({W}^{(\ell)} \overline{{s}}^{(\ell)}, T, V_{t h}^{(\ell)}\right) \nonumber \\
&= \frac{V_{t h}^{(\ell)}}{T} {Clip}\left(\left\lfloor\frac{T}{V_{t h}^{(\ell)}} {W}^{(\ell)} \overline{{s}}^{(\ell)}\right\rfloor, 0, T\right) 
\end{align}
\noindent where $\lfloor x \rfloor$ refers to the round down operator. The $Clip$ function limits above but allows below. Although these methods are promising, they often require extensive re-training epochs (hundreds of hours, as shown in Tab~\ref{tb:comparison}) to achieve optimal weights and thresholds, as well as prolonged timesteps during inference, making them computationally intensive.

To address the retraining burden, Li et al.~\cite{liFreeLunchANN2021} proposed a layer-wise Calibration algorithm designed to minimize the discrepancy between the output of original ANNs and the converted SNNs. This Spike Calibration method determines the optimal threshold by leveraging Eq.~\ref{eq3}:
\begin{equation}
\label{eq4}
\min _{V_{t h}^{(\ell)}}\left(ClipFloor\left(\overline{{s}}^{(\ell+1)}, T, V_{t h}^{(\ell)}\right)-ReLU\left(\overline{{s}}^{(\ell+1)}\right)\right)^2
\end{equation}

However, previous studies overlook the energy efficiency of converted SNNs and require times of inference timesteps to achieve optimal performance than directly trained SNNs.


\begin{figure*}[!t]
\centering
\includegraphics[width=0.95\linewidth]{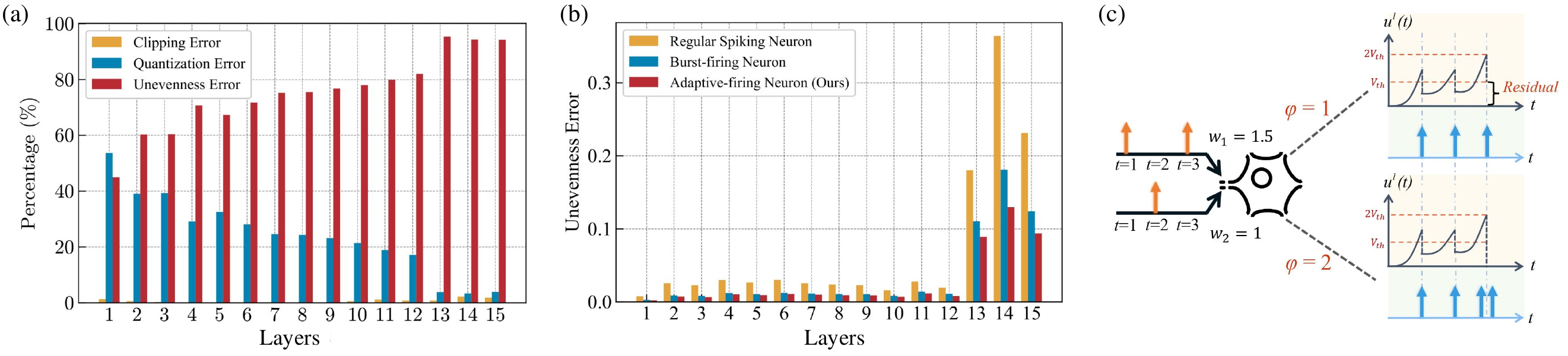}
\vskip -0.05in
\caption{{\textbf{The unevenness error dominates conversion loss in ANN-to-SNN conversion.} (a) Percentage of three main conversion errors, with the unevenness error dominating. (b) The adoption of Adaptive-firing Neurons greatly reduce the unevenness error. (c) Burst-firing mechanism in the Adaptive-firing Neuron model. The Adaptive-firing Neuron model minimizes this loss by allowing multiple spikes to be generated in rapid succession when the membrane potential exceeds the threshold.}}
\vskip -0.1in
\label{fig:bs}
\end{figure*}

\section{Adaptive Calibration}
\subsection{Motivation}
 The ANN-to-SNN conversion process introduces three main types of errors: clipping error, quantization error, and unevenness error~\cite{buOptimalANNSNNConversion2021, hao2023reducing, haobridging}. As illustrated in Fig.~\ref{fig:bs}(a), the unevenness error is the dominant factor during conversion. Therefore, the key challenge lies in reducing this error within limited timesteps.

\noindent\textbf{Unevenness Error.} The unevenness error is defined as the difference between the average output of the converted SNN and the output of the source ANN~\cite{hao2023reducing}:
\begin{equation}
E^{(\ell)}  = \overline{s}^{(\ell)} - x^{(\ell)} 
\label{eq:uneven}
\end{equation}
This error quantifies the discrepancy between the expected and actual output of the converted SNN. It primarily occurs when input spikes are unevenly distributed, particularly when high-frequency inputs exceed a neuron's firing capabilities within limited timesteps. Fig.~\ref{fig:bs}(c) illustrates this concept: consider a neuron receiving three spikes (two weighing 1.5 each and one weighing 1). In an ANN, the neuron's expected output would be 4. However, in an SNN constrained to fire only once per timestep over three timesteps, the actual activation achieves only 3, resulting in a 1-unit unevenness error. This error manifests as residual membrane potential in the neuron, which our Adaptive Calibration method aims to address.

\noindent\textbf{Bio-inspired Burst-Firing Neuron.} To mitigate this error and enhance information processing efficiency, we can turn to biological systems for inspiration. In the brain, brief bursts of high-frequency firing play a crucial role in enhancing neural communication reliability. This diversity in neuronal responses allows the neocortex to dynamically adjust its information processing based on input characteristics and network demands, optimizing both performance and efficiency~\cite{connorsIntrinsicFiringPatterns1990, izhikevichBurstsUnitNeural2003, lismanBurstsUnitNeural1997}. Fig.\ref{fig:bs}(b) demonstrates that bio-inspired burst-firing neurons can significantly reduce the unevenness error. This reduction occurs because burst-firing neurons have more diverse firing patterns and an increased capacity to handle uneven input spikes. Fig.\ref{fig:bs}(c) exemplifies how allowing a maximum firing time $\varphi$ of 2 can reduce unevenness and mitigate conversion loss. The burst-firing neuron model can expand the potential range of neuronal activation output $\overline{{s}}^{(\ell)}$ to \([0, V_{t h}^{(\ell-1) } \times \varphi] \). Consequently, the relationship between the activation output of ANNs and converted SNNs (Eq.~\ref{eq4}) becomes:
\begin{align}
\label{eq6}
\overline{{s}}^{(\ell+1)} &= ClipFloor\left({W}^{(\ell)} \overline{{s}}^{(\ell)}, T, V_{t h}^{(\ell)}, \varphi^{(\ell)}\right) \nonumber \\
&= \frac{V_{t h}^{(\ell)}}{T} Clip\left(\left\lfloor\frac{T}{V_{t h}^{(\ell)}} {W}^{(\ell)} \overline{{s}}^{(\ell)}\right\rfloor, 0, T \times \varphi\right)
\end{align}
While burst-firing neurons can reduce the unevenness error by increasing the neuron's capacity for rapid firing, this approach also allows for the generation of a large number of spikes, potentially leading to significant energy consumption. To address this challenge, we propose an adaptive calibration framework that provides a unified solution to significantly reduces the unevenness error while simultaneously decreasing energy consumption and latency.

\subsection{Preliminary}
\noindent\textbf{Metrics for Performance and Efficiency}
The success of our Adaptive Calibration depends on precise metrics reflecting the performance and efficiency of the converted SNNs.

\noindent (1) Performance Metric: We estimate the performance of SNNs using sensitivity, which is demonstrated inversely related to SNN performance (shown in the Appendix). To quantify layer sensitivity to a parameter $k$, we employ Kullback-Leibler (KL) divergence~\cite{caiZeroqNovelZero2020}:
\begin{equation}
\label{eq8}
S_i(k) = \frac{1}{N} \sum_{j=1}^{N} \mathrm{KL}\left(\mathcal{M}\left(\mathrm{ANN_i}; x_j\right), \mathcal{M}\left(\mathrm{SNN_i(k)} ; x_j\right)\right)
\end{equation}
This sensitivity metric for layer $i$ relative to parameter $k$ measures the difference in output distributions between the ANN and SNN configurations. A lower $S_i(k)$ indicates closer alignment of the SNN's output with the ANN's, reflecting reduced sensitivity and potentially higher performance.

\noindent (2) Efficiency Metric: For assessing SNNs' efficiency and energy consumption, we draw upon established methodologies~\cite{wang2023masked, dingOptimalAnnsnnConversion2021, cao2015spiking}, calculating energy based on the total number of spikes and their associated energy cost per spike, \( \mu \) Joules:
\begin{equation}
   E = \frac{\text{total spikes}}{1 \times 10^{-3}} \times \mu \quad (\text{in Watts})
\label{eq9}
\end{equation}


\begin{figure*}[!t]
\centering
\includegraphics[width=0.9\linewidth]{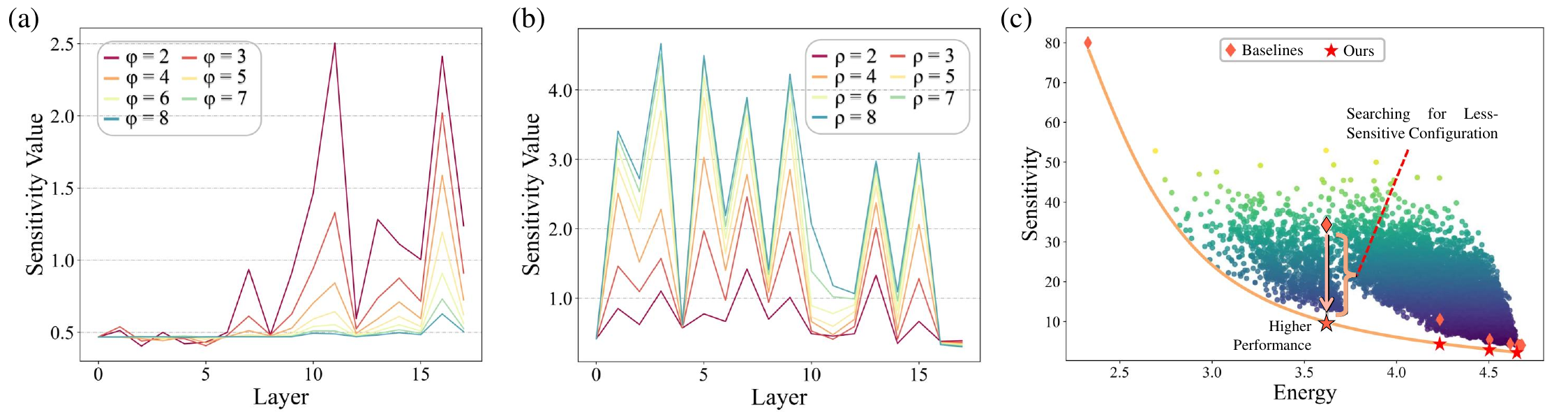}
\vskip -0.05in
\caption{(a-b) \textbf{Variation of the sensitivity of each layer} with respect to: (a) different max burst-firing patterns $\varphi$ and (b) different threshold ratios $\rho$. (c) \textbf{Pareto Frontier Searching.} Optimizing the network configuration to reduce sensitivity improves performance, where each data point represents a distinct layer-specific configuration. }
\vskip -0.1in
\label{fig:vis}
\end{figure*}


\subsection{Adaptive-Firing Neuron Model} \label{AdaFire}
\noindent\textbf{Observation 1:} \textit{The sensitivity to variations in max burst-firing pattern 
 \(\varphi\) differs significantly across network layers.}

\noindent This observation, shown in Fig~\ref{fig:vis}(a), serves as a cornerstone for our novel approach. Building on this observation, we propose that layers with higher sensitivity to \(\varphi\) variations should be given a broader range of firing patterns, and vice versa. This leads to the development of the \textbf{Adaptive-Firing Neuron Model (AdaFire)}, which strategically accounts for each layer's sensitivity to firing pattern changes while also aiming to minimize energy consumption.

\noindent\textbf{Layer-Specific Firing Patterns Adaptation.}
Applying uniform burst-firing capabilities, denoted as \( \varphi \), across all layers in SNNs may not be the most effective approach. This idea is motivated by the observation that biological neurons exhibit diverse firing patterns, which are adapted to their particular functional roles within the neural network~\cite{connorsIntrinsicFiringPatterns1990, izhikevichBurstsUnitNeural2003, lismanBurstsUnitNeural1997}. Our observation, depicted in Fig.~\ref{fig:vis}(a), also demonstrates this significant insight. Therefore, layer-specific \(\varphi\) is crucial. 

\noindent\textbf{Pareto Frontier Driven Search Algorithm.}
Optimizing the layer-specific \( \varphi \) is a non-trivial problem. For an SNN model with \( L \) layers and \( n \) configurations per layer, the solution space is \( n^L \), growing exponentially with more layers. To tackle this complexity, inspired by~\cite{caiZeroqNovelZero2020}, we introduce a simplifying assumption: each layer's sensitivity to its configuration is independent of other layers' configurations. This assumption allows us to decompose the global optimization problem into a series of local, layer-wise optimizations. Consequently, we can reduce the search space from $O(n^L)$ to $O(nL)$, making the problem tractable for practical network sizes. Under this framework, we formulate our objective to identify the optimum combination that minimizes the overall sensitivity value $S_{\text{sum}}$ within a predefined energy budget $E_{\text{target}}$. Leveraging the Pareto frontier approach, we seek configurations that balance sensitivity reduction and energy consumption. Formally, the optimization problem is defined as:
\begin{equation}
\label{eq10}
\min _{\left\{\varphi_i\right\}_{i=1}^L} S_{\text{sum}} = \sum_{i=1}^L S_i\left(\varphi_i\right), \quad\sum_{i=1}^L E_i \leq E_{\text{target}}
\end{equation}
where $\varphi_i$ represents the chosen configuration for the $i^{th}$ layer, and $E_i$ denotes the estimated energy consumption for that layer. This formulation allows us to optimize performance as a sum of individual layer sensitivities, significantly simplifying the search process. As shown in Fig.~\ref{fig:vis}(c), our method effectively balances the trade-off between energy consumption and sensitivity, outperforming baseline approaches that lack a systematic search strategy.

\subsection{Sensitivity Spike Compression}

While the Adaptive-firing Neuron model can automatically search for optimum burst-firing patterns for each layer to improve performance within a defined energy budget, it also allows for increased spike generation, potentially leading to significant energy consumption. To address this, we focus on reducing energy consumption during the conversion process, an often-neglected aspect of ANN-to-SNN conversion.

\noindent\textbf{Observation 2:} \textit{ The sensitivity to variations in threshold ratio 
\(\rho\) varies distinctly across network layers.}

\noindent Leveraging this insight, we optimize network efficiency through strategic spike inhibition. By assigning higher \(\rho\) values to less sensitive layers, we achieve a substantial reduction in overall spiking activity of the network, as shown in Fig.~\ref{fig:vis}(b).  


\noindent\textbf{Adaptive Threshold.}
As depicted in Fig.~\ref{fig:thres}, when a neuron emits spikes at regular intervals, consecutive spikes can be compressed by a singular, double-amplitude spike without losing the original timing information. This process can be mathematically represented as:
\begin{equation}
V_{t h}^{(\ell)}=\rho^{(\ell)} \cdot v_{t h}^{(\ell)}
\label{eq11}
\end{equation}
where \(\rho^{(\ell)} \) refers to the threshold amplification ratio and \( v_{t h}^{(\ell)} \) signifies the initial threshold of layer \( \ell \). The subsequent spike output of an IF neuron can be described by:
\begin{align}
\label{eq12}
{s}^{(\ell+1)}(t) &= \begin{cases} 
\rho^{(\ell)} \cdot V_{t h}^{(\ell)} & \text{if } {u}^{(\ell)}(t+1) \geq \rho^{(\ell)} \cdot V_{t h}^{(\ell)} \\
0 & otherwise
\end{cases}
\end{align}
Subsequently, the updated firing rate for SNN output is:
\begin{equation}
{r}^{(\ell+1)} = \sum_{i=1}^n {W}_i^{(\ell)} \frac{\sum_{t=1}^T {s}_i^{(\ell)}(t) \cdot \rho^{(\ell)}}{T}
\label{eq13}
\end{equation}
This method effectively decreases the spike generation while ensuring that the quantity of information conveyed through each neuron is amplified by the factor \(\rho^{(\ell)}\), maintaining the integrity of information transmission across layers.

{
\centering
\includegraphics[width=0.85\linewidth]{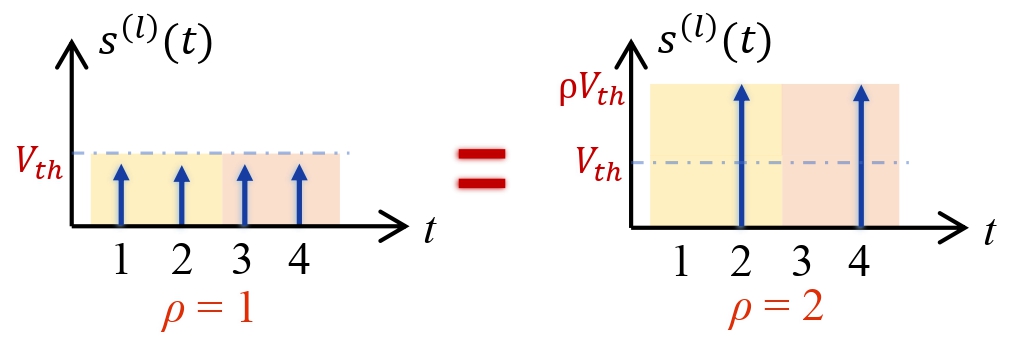}
\captionof{figure}{\textbf{Spike Compression Mechanism.} Our approach enables the compression of regular spikes.}
\label{fig:thres}
\vskip 0.05in}

\noindent\textbf{Adaptive Threshold Search Algorithm.}
Naively applying threshold compression can significantly degrade performance, especially with irregular spike trains where compression could lead to data loss. 
To mitigate this, we propose the Sensitivity Spike Compression (SSC) method. SSC assesses how changes in the threshold ratio \(\rho\) affect output variability. For each layer, the goal is to determine the optimum \(\rho\) that minimizes spike generation while maintaining accuracy. Building on the key insight from Observation 2, we formulate the goal as follows:
\begin{equation}
\min_{\{\rho_i\}_{i=1}^L} E_{\text{sum}} = \sum_{i=1}^L E_i(\rho_i), \quad \sum_{i=1}^L S_i \leq S_{\text{target}}.
\label{eq14}
\end{equation}

\subsection{Input-aware Adaptive Timesteps}
To further enhance the efficiency of converted SNNs, we focus on optimizing the timesteps $T$. Traditionally, $T$ is set as a fixed hyperparameter in SNN configurations. However, this static approach fails to capitalize on the potential benefits of dynamically adjusting timesteps to accommodate the unique characteristics of each input image. Recent studies have highlighted the capacity of SNNs to adapt timesteps dynamically based on individual input features ~\cite{li2024seenn}.

\noindent\textbf{Entropy as a Confidence Measure.}
Inspired by~\cite{teerapittayanon2016branchynet, guo2017calibration}, we employ entropy as a confidence measure for predictions at each timestep. Formally, the entropy \(H(p)\) is defined as:
\begin{equation}
    H(p) = \sum_{y \in \mathcal{Y}} p_y \log p_y,
    \label{eq15}
\end{equation}
where \(p_y\) represents the probability of label \(y\). 

\noindent\textbf{Dynamic Timestep Adjustment Mechanism.}
We adopt a confidence level-based mechanism, with a predefined boundary \(\alpha\), to dynamically determine the required inference timestep. During the inference process, our SNN exits as soon as the confidence score exceeds the boundary \(\alpha\), thus optimizing the balance between accuracy and latency. Our research reveals a crucial empirical finding (shown in the Appendix): each timestep contributes differently to the network's final accuracy. This insight distinguishes our approach from previous works that rely on a single fixed boundary $\alpha$ for all timesteps~\cite{li2024seenn}. Consequently, we propose a dynamic threshold for the confidence score at each timestep, defined as:
\begin{equation}
\label{eq18}
\alpha_{t}=\alpha_{base}+\beta e^{-\frac{\bar{E}_t-\bar{E}_{min}}{\delta}}
\end{equation}
where $\alpha_{base}$ is the base boundary, $\beta$ is the scaling factor, $\delta$ represents the decay constant, $\bar{E}_t$ denotes the average entropy of the network's output distribution associated with each timestep $t$, and $\bar{E}_{min}$ is the minimum value within $\bar{E}_t$. This formulation allows for a dynamic boundary \(\alpha\) adjustment:  higher average entropy at a given timestep, indicating lower confidence in the output, warrants a higher simulation time to ensure accurate inference. The detailed methodology and pseudo-code are provided in the Appendix.

\begin{table}[!t]
\centering
\caption{Performance comparison between the proposed model and the state-of-the-art models on the ImageNet dataset. Rt. represents the need for re-training.}
\vskip -0.05in
\label{tb:imagenet}
\scriptsize
\renewcommand{\arraystretch}{1}
\setlength\tabcolsep{2pt}
\begin{tabular}{llcccccc}
\toprule
\textbf{Arch.} & \textbf{Method} & \textbf{ANN} & \textbf{T=8} & \textbf{T=16} & \textbf{T=32} & \textbf{T=64} & \textbf{Rt.} \\
\midrule
\multirow{6}{*}{\textbf{VGG-16}} 
& OPT~\cite{dengOptimalConversionConventional2021}\textsuperscript{ICLR} & 75.36 & - & - & 0.11 & 0.12 & \cmark \\
& SNM~\cite{wangSignedNeuronMemory2022}\textsuperscript{IJCAI} & 73.18 & - & - & 64.78 & 71.50 & \cmark \\
& QCFS~\cite{buOptimalANNSNNConversion2021}\textsuperscript{ICLR} & 74.39 & - & 50.97 & 68.47 & 72.85 & \cmark \\
& SRP~\cite{hao2023reducing}\textsuperscript{AAAI} & 74.29 & 68.37 & 69.13 & 69.35 & 69.43 & \cmark \\
& Calibration~\cite{liFreeLunchANN2021}\textsuperscript{ICML} & 75.36 & 25.33 & 43.99 & 62.14 & 65.56 & \xmark \\
& \cellcolor{mygray}\textbf{AdaFire (Ours)} & \cellcolor{mygray}75.36 & \cellcolor{mygray}73.53 & \cellcolor{mygray}74.25 & \cellcolor{mygray}74.98 & \cellcolor{mygray}75.22 & \cellcolor{mygray}\xmark \\
\midrule
\multirow{5}{*}{\textbf{ResNet-34}} 
& OPT~\cite{dengOptimalConversionConventional2021}\textsuperscript{ICLR} & 75.66 & - & - & 0.11 & 0.12 & \cmark \\
& QCFS~\cite{buOptimalANNSNNConversion2021}\textsuperscript{ICLR} & 74.32 & - & - & 69.37 & 72.35 & \cmark \\
& SRP~\cite{hao2023reducing}\textsuperscript{AAAI} & 74.23 & 67.62 & 68.02 & 68.40 & 68.61 & \cmark \\
& Calibration~\cite{liFreeLunchANN2021}\textsuperscript{ICML} & 75.66 & 0.25 & 34.91 & 61.43 & 69.53 & \xmark \\
& \cellcolor{mygray}\textbf{AdaFire (Ours)} & \cellcolor{mygray}75.66 & \cellcolor{mygray}72.96 & \cellcolor{mygray}73.85 & \cellcolor{mygray}75.04 & \cellcolor{mygray}75.38 & \cellcolor{mygray}\xmark \\
\midrule
\multirow{2}{*}{\textbf{ViT}} 
& Calibration~\cite{liFreeLunchANN2021}\textsuperscript{ICML} & 79.36 & 0.34 & 3.58 & 38.36 & 60.45 & \xmark \\
& \cellcolor{mygray}\textbf{AdaFire (Ours)} & \cellcolor{mygray}79.36 & \cellcolor{mygray}\textbf{68.08} & \cellcolor{mygray}\textbf{74.22} & \cellcolor{mygray}\textbf{76.36} & \cellcolor{mygray}\textbf{77.09} & \cellcolor{mygray}\xmark \\
\bottomrule 
\end{tabular}
\vskip -0.1in
\end{table}

\section{Experiment}
We conducted a comprehensive evaluation of our Adaptive Calibration framework across a diverse range of benchmarks, including tasks in 2D and 3D classification, event-driven classification, object detection, and segmentation. As illustrated in Fig~\ref{fig:radar}, our framework demonstrates superior performance across all tasks. Detailed results and additional experiments (3D tasks and segmentation) are provided in the Appendix.
\vskip -0.05in

\subsection{Effectiveness of Adaptive-Firing Neuron Model}

\noindent\textbf{Performance on Static Classification.}
We evaluated the Adaptive-firing Neuron Model (AdaFire) using the ImageNet dataset \cite{dengImagenetLargescaleHierarchical2009}. Tab.~\ref{tb:imagenet} shows that AdaFire maintains high accuracy with fewer timesteps compared to leading conversion methods. Notably, our method achieves these results \textbf{without requiring additional training}. As shown in Tab.~\ref{tb:comparison}, our method requires only 0.09 hours of setup time, in stark contrast to methods like QCFS~\cite{buOptimalANNSNNConversion2021}, which demands 742 hours for additional training. This significant reduction in setup time translates to improved practicality and faster deployment. For a fair comparison, we evaluated our model at $T=8$ against competitors at $T=32$, equalizing energy consumption by setting $\varphi$ to 4. Results indicate that AdaFire exceeds the Calibration base framework by \textbf{11.39\%}, and outperforms QCFS~\cite{buOptimalANNSNNConversion2021} and SNM~\cite{wangSignedNeuronMemory2022} by \textbf{5.06\%} and \textbf{8.75\%}, respectively, on the VGG-16 architecture. In addition, we further evaluate the versatility of our method on other advanced models, like Vision Transformer~\cite{dosovitskiy2020image}. The results show that our model can also achieve better performance within a limited timestep, underscoring its broad applicability.

\noindent\textbf{Performance on Event-driven Classification.}
Tab.~\ref{tb:neuromorphic} presents our evaluation across various neuromorphic datasets, such as CIFAR10-DVS and N-Caltech101, which were derived from static datasets using event-based cameras. AdaFire consistently outperformed leading SNN models, including PLIF~\cite{fangIncorporatingLearnableMembrane2021}, by \textbf{6.45\%} using only 8 timesteps. More results are shown in the Appendix.

\begin{table}[!t]
\centering
\scriptsize
\setlength\tabcolsep{9pt}
\caption{Performance comparison between the proposed model and the state-of-the-art models on different neuromorphic datasets.}
\vskip -0.05in
\label{tb:neuromorphic}
\begin{tabular}{llcc}
    \toprule 
    \textbf{Dataset} & \textbf{Model} & \textbf{T} & \textbf{Acc. (\%)} \\
    \midrule
    \multirow{4}{*}{\textbf{CIFAR10-DVS}} & PLIF~\cite{fangIncorporatingLearnableMembrane2021}\textsuperscript{ICCV} & 20 & 74.80 \\
    & Dspkie~\cite{liDifferentiableSpikeRethinking2021}\textsuperscript{NeurIPS} & 10 & 75.40 \\
    & DSR ~\cite{mengTrainingHighPerformanceLowLatency2022}\textsuperscript{CVPR} & 10 & 77.30 \\
    & \cellcolor{gray!25}\textbf{AdaFire (Ours)} & \cellcolor{gray!25}\textbf{8} & \cellcolor{gray!25}\textbf{81.25} \\
    \midrule
    \multirow{3}{*}{\textbf{N-Caltech101}} & SALT~\cite{kimOptimizingDeeperSpiking2021a}\textsuperscript{NN} & 20 & 55.00 \\
    & NDA~\cite{liNeuromorphicDataAugmentation2022}\textsuperscript{ECCV} & 10 & 83.70 \\
    & \cellcolor{gray!25}\textbf{AdaFire (Ours)} & \cellcolor{gray!25}\textbf{8} & \cellcolor{gray!25}\textbf{85.21} \\
    \bottomrule  
  
\end{tabular}
  \vspace{-0.05in}
\end{table}

\begin{table}[!t]
\centering
\scriptsize
\setlength\tabcolsep{5.5pt}
\caption{Performance comparison for object detection on PASCAL VOC 2012 and MS COCO 2017 datasets. mAP represents the mean Average Precision.}
\label{tab:detection}
\vskip -0.05in
\begin{tabular}{llccc}
    \toprule
    \textbf{Dataset} & \textbf{Method}  & \textbf{ANN} & \textbf{T} & \textbf{mAP} \\
    \midrule
    \multirow{4}{*}{\textbf{VOC}} & Spiking-YOLO~\cite{kimSpikingyoloSpikingNeural2020}\textsuperscript{AAAI}  & 53.01 & 8000 & 51.83 \\
    & B-Spiking-YOLO~\cite{kimFastAccurateObject2020}\textsuperscript{Access} & 53.01 & 5000 & 51.44 \\
    & Calibration~\cite{liFreeLunchANN2021}\textsuperscript{ICML}  & 54.34 & 128 & 47.15 \\
    & \cellcolor{gray!25}\textbf{AdaFire (Ours)}  & \cellcolor{gray!25}54.34 & \cellcolor{gray!25}16 & \cellcolor{gray!25}\textbf{51.91} \\
    \midrule
    \multirow{4}{*}{\textbf{COCO}} & Spiking-YOLO~\cite{kimSpikingyoloSpikingNeural2020}\textsuperscript{AAAI} &  26.24 & 8000 & 25.66 \\
    & B-Spiking-YOLO~\cite{kimFastAccurateObject2020}\textsuperscript{Access} & 26.24 & 5000 & 25.78 \\
    & Calibration~\cite{liFreeLunchANN2021}\textsuperscript{ICML}& 26.78 & 128 & 20.12 \\
    & \cellcolor{gray!25}\textbf{AdaFire (Ours)}  & \cellcolor{gray!25}26.78 & \cellcolor{gray!25}16 & \cellcolor{gray!25}\textbf{26.13} \\
    \bottomrule
\end{tabular}
\vspace{-0.1in}
\end{table}

\noindent\textbf{Performance on Object Detection.}
We evaluated our method on the PASCAL VOC 2012 and MS COCO 2017 datasets, benchmarked against established models on the Tiny-YOLO model. Tab.~\ref{tab:detection} shows our method's substantial efficiency improvement on COCO, where it achieved a mAP of \textbf{26.13\%} with only 16 timesteps, compared to Spiking-YOLO's~\cite{kimSpikingyoloSpikingNeural2020} 25.66\% mAP at 8000 timesteps. This represents a \textbf{500$\times$} speed-up, highlighting our method's potential for real-time applications.

\begin{figure*}[!t]
    \centering
    \begin{subfigure}[b]{0.3\textwidth}
        \centering
        \includegraphics[width=\textwidth]{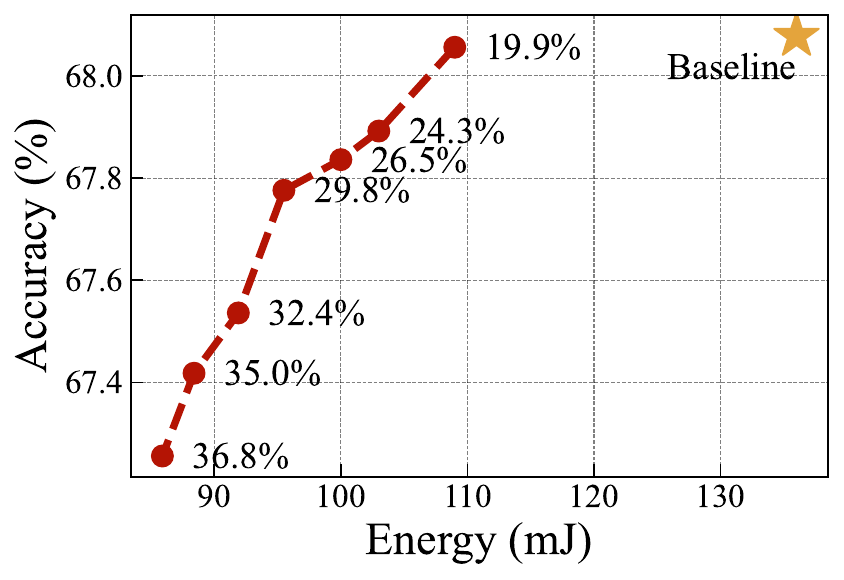} 
        \vskip -0.1in
        \caption{ImageNet}
        \label{fig:imagenet}
    \end{subfigure}
    \begin{subfigure}[b]{0.3\textwidth}
        \centering
        \includegraphics[width=\textwidth]{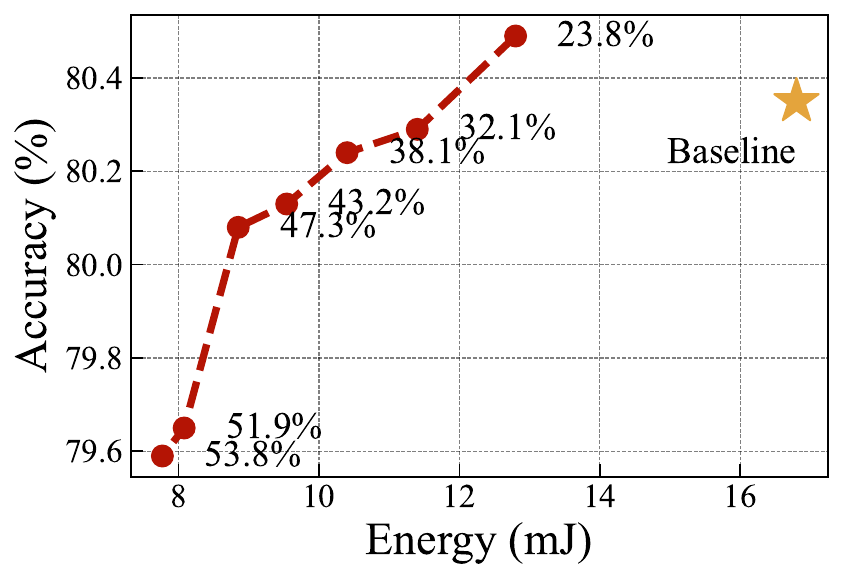} 
        \vskip -0.1in
        \caption{CIFAR-100}
        \label{fig:cifar100}
    \end{subfigure}
    \begin{subfigure}[b]{0.3\textwidth}
        \centering
        \includegraphics[width=\textwidth]{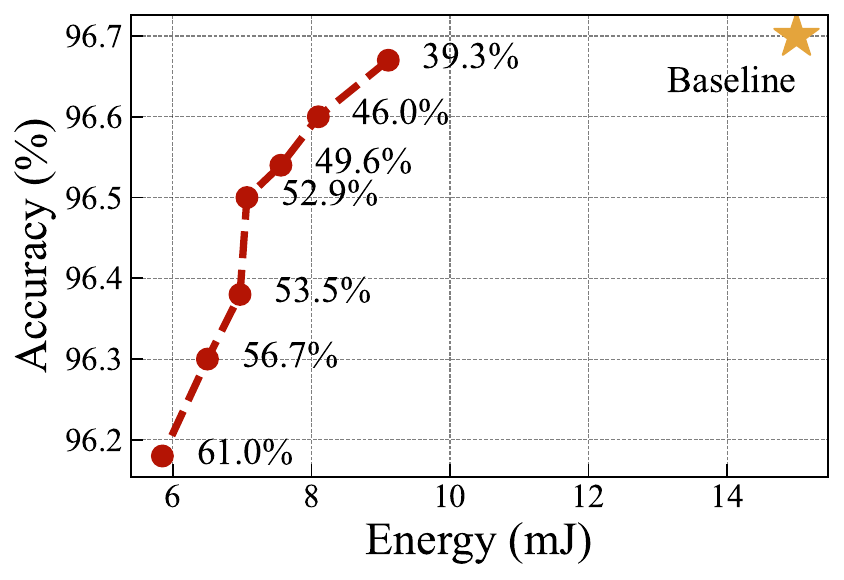} 
        \vskip -0.1in
        \caption{CIFAR-10}
        \label{fig:cifar10}
    \end{subfigure}
    \vskip -0.1in
    \caption{\textbf{Effectiveness of Sensitivity Spike Compression (SSC)}. The baseline is the results without using the SSC.}
    \vskip -0.1in
    \label{fig:SSC}
\end{figure*}


\begin{figure*}[!t]
\centering
\includegraphics[width=0.95\linewidth]{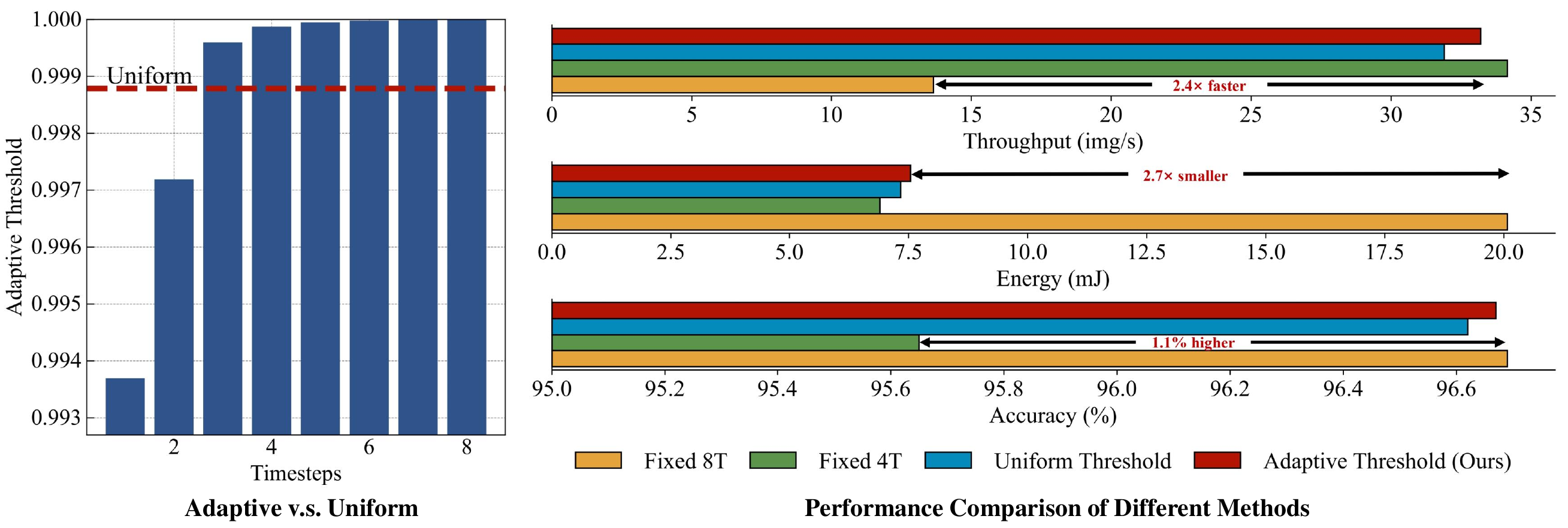}
\vskip -0.05in
\caption{\textbf{Effectiveness of Input-aware Adaptive Timesteps Technique.}}
\vskip 0.1in
\label{fig:comprarision}
\end{figure*}

\begin{table*}[t!]
\renewcommand{\arraystretch}{1}
    \setlength\tabcolsep{10pt} 
    \centering
\scriptsize
    \caption{Ablation Study of Different Techniques of our Adaptive Calibration. $T$ is set to 8 by default.}
    \vskip -0.1in
    \label{tab4}
\begin{tabular}{ccc|cc|cc|cc}
\hline
\multirow{2}{*}{\textbf{AdaFire}} & \multirow{2}{*}{\textbf{SSC}} & \multirow{2}{*}{\textbf{IAT}} & \multicolumn{2}{c}{\textbf{CIFAR-10 (ResNet-20)}}         & \multicolumn{2}{c}{\textbf{CIFAR-100 (ResNet-20)}}        & \multicolumn{2}{c}{\textbf{ImageNet (ResNet-18)}}         \\
                                  &                               &                               & \textbf{Accuracy (\%)} & \textbf{Energy (mJ)} & \textbf{Accuracy (\%)} & \textbf{Energy (mJ)} & \textbf{Accuracy (\%)} & \textbf{Energy (mJ)} \\ \hline
                                  &                               &                               & 96.34                  & 14.86  (0)           & 79.90                   & 16.83 (0)            & 56.74                  & 162.56 (0)               \\
\cmark                                 &                               &                               & \textbf{96.69}                  & 20.49 (+37.88\%)    & \textbf{80.64}                  & 21.44 (+27.37\%)     & \textbf{68.45}                  & 169.52 (+0.04\%)     \\
\cmark                                 & \cmark                             &                               & 96.5                   & 7.71 (-48.12\%)      & 80.37                  & 10.00 (-40.58\%)        & 68.32                  & 120.32 (-25.98\%)       \\
\cmark                                 &                               & \cmark                             & 96.67                  & 7.06 (-52.48\%)      & 80.55                  & 12.16 (-27.75\%))    &       68.39                 &  105.26 (-35.25\%)                   \\
\cmark                                 & \cmark                             & \cmark                             & 95.47                  & \textbf{4.44 (-70.12\%)}      & 80.00                     & \textbf{6.69 (-60.25\%))}     &  68.27                      &     \textbf{92.50 (-43.10\%)}                
    \\\hline
    \end{tabular}
    \vskip -0.1in
\end{table*}

\subsection{Effectiveness of Sensitivity Spike Compression}
We evaluate the Sensitivity Spike Compression (SSC) technique on CIFAR-10, CIFAR-100, and ImageNet datasets, tuning \( S_{target} \) as per Eq~\ref{eq14} to balance energy consumption and performance. The method of calculating the theoretical energy is shown in the Appendix. Results in Fig.~\ref{fig:SSC} show a \textbf{61.0\%} energy reduction on CIFAR-10 with minimal accuracy loss (0.5\%). On ImageNet, SSC achieved a \textbf{32.4\%} energy saving with a comparable accuracy decrease. These results validate SSC's efficiency in enhancing the energy economy without significant performance trade-offs.

\subsection{Effectiveness of Input-aware Adaptive Timesteps}
As illustrated in Fig.~\ref{fig:comprarision}(a), our Input-aware Adaptive Timesteps (IAT) method dynamically adjusts the confidence threshold $\alpha$ to optimize processing time. This approach reduces latency and increases energy efficiency by allowing early exits for simpler images and extended processing for complex ones. Fig.~\ref{fig:comprarision}(b) shows a \textbf{2.4-fold} increase in speed and a \textbf{2.7-fold} reduction in energy consumption, with a performance improvement of \textbf{1.1\%} over the baseline. These outcomes highlight the IAT technique's effectiveness in lowering latency and energy costs without compromising accuracy.

\vskip -0.05in
\subsection{Ablation Study}

We evaluate three proposed techniques—AdaFire, SSC, and IAT on the CIFAR-10, CIFAR-100, and ImageNet datasets. Our results reveal that the AdaFire significantly boosts the accuracy of SNNs. Concurrently, the SSC and IAT techniques contribute to a substantial reduction in energy consumption. Remarkably, the synergistic application of three techniques leads to a groundbreaking \textbf{70.12\%} energy reduction and a \textbf{0.13\%} accuracy enhancement for the CIFAR-10 dataset. For the more challenging ImageNet dataset, the combined implementation achieves a \textbf{43.10\%} decrease in energy consumption while simultaneously enhancing accuracy by \textbf{11.53\%}. These results underscore the efficacy of our proposed conversion framework as a unified solution capable of both improving performance and efficiency.

\section{Conclusion}
In our paper, we propose a unified training-free ANN-to-SNN conversion framework optimized for both performance and efficiency. We introduce the Adaptive-firing Neuron Model (AdaFire), which automatically searches for optimum burst-firing patterns of each layer, significantly improving the SNN performance at low timesteps. Moreover, to improve efficiency, we propose a Sensitivity Spike Compression (SSC) technique and an Input-aware Adaptive Timesteps (IAT) technique, reducing both the energy consumption and latency during the conversion process.

\section{Acknowledgments}
This work is supported by the Guangzhou-HKUST(GZ) Joint Funding Program (Grant No. 2023A03J0682) and partially supported by the collaborative project with Brain Mind Innovation, inc.


\bibliography{main}

\end{document}